\documentclass[sigconf]{acmart}
\usepackage{enumitem}
\usepackage{multirow}
\usepackage{amsmath}
\usepackage{xcolor}
\usepackage{colortbl}
\usepackage{arydshln}
\usepackage{utfsym}
\usepackage{fontawesome}
\usepackage{listings}
\title{Visual Room 2.0: Seeing is Not Understanding for MLLMs}

\author{Haokun Li}
\affiliation{
  \institution{Tianjin University}
  \country{China}
}
\email{lihaokun@tju.edu.cn}

\author{Yazhou Zhang}
\authornote{Corresponding author.}
\affiliation{
  \institution{Tianjin University}
  \country{China}
}
\email{yzhou_zhang@tju.edu.cn}

\author{Jizhi Ding}
\affiliation{
  \institution{Shandong Institute of Petroleum and Chemical Technology}
  \country{China}
}
\email{dingjizhi@sdipct.edu.cn}

\author{Qiuchi Li}
\affiliation{
  \institution{Beijing Institute of Technology}
  \country{China}
}
\email{liqiuchi2015@gmail.com}
\author{Peng Zhang}
\authornotemark[1]
\affiliation{
  \institution{Tianjin University}
  \country{China}
}
\email{pzhang@tju.edu.cn}

\acmConference[WWW'26 Short Papers]{The Web Conference 2026 (Short Papers Track)}{April 27–May 1, 2026}{Dubai}
\acmBooktitle{Proceedings of the ACM Web Conference 2026 (WWW '26 Short Papers), April 27–May 1, 2026, Singapore}
\acmPrice{15.00}
\acmISBN{978-1-4503-XXXX-X/26/04}
\acmDOI{10.1145/nnnnnnn.nnnnnnn}

\begin{document}

\begin{abstract}
Can multi-modal large language models (MLLMs) truly understand what they can see?
Extending Searle's \textit{Chinese Room} into the multi-modal domain, this paper proposes the \textit{Visual Room} argument: MLLMs may describe every visual detail precisely yet fail to comprehend the underlying emotions and intentions, namely seeing is not understanding.
Building on this, we introduce \textit{Visual Room} 2.0, a hierarchical benchmark for evaluating perception–cognition alignment of MLLMs.
We model human perceptive and cognitive processes across three levels: low, middle, and high, covering 17 representative tasks.
The perception component ranges from attribute recognition to scene understanding, while the cognition component extends from textual entailment to causal and social reasoning.
The dataset contains 350 multi-modal samples, each with six progressive questions (2,100 in total) spanning perception to cognition.
Evaluating 10 state-of-the-art (SoTA) MLLMs, we highlight three key findings:
(1) MLLMs exhibit stronger perceptual competence than cognitive ability (8.0\%$\uparrow$);
(2) cognition appears not causally dependent on perception-based reasoning; and
(3) cognition scales with model size, but perception does not consistently improve with larger variants.
This work operationalizes ``Seeing $\ne$ Understanding'' as a testable hypothesis, offering a new paradigm from perceptual processing to cognitive reasoning in MLLMs\footnote{Our dataset is available at \url{https://huggingface.co/datasets/LHK2003/PCBench}}.
\end{abstract}



\maketitle
\begin{figure}[t!]
    \centering
    \includegraphics[width=0.50\textwidth]{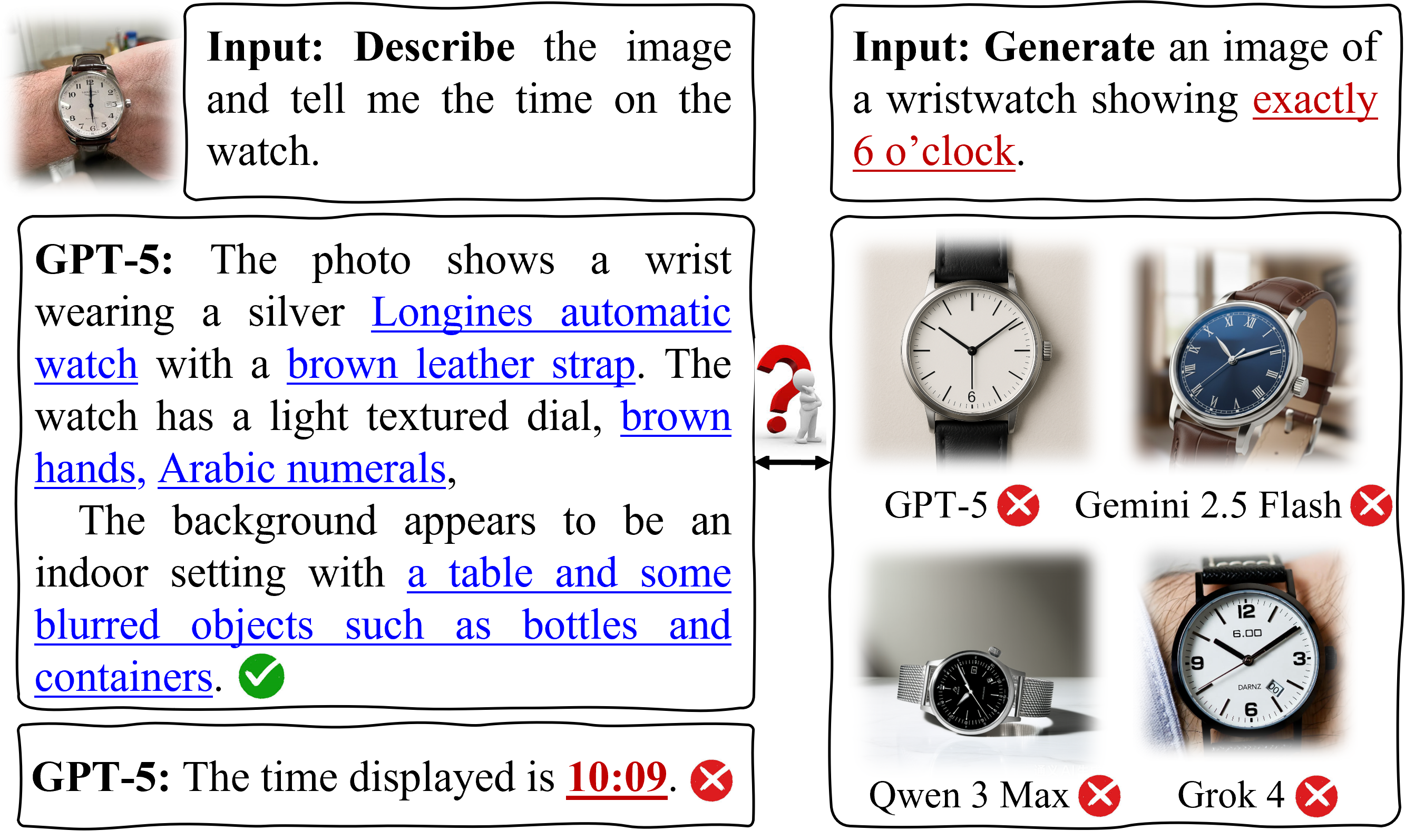}
    \caption{The watch problem.}
    \label{fig:watch}
\end{figure}
\section{Introduction}
MLLMs have demonstrated impressive performance on a broad set of visual downstream tasks, e.g., visual question answering, object detection, image captioning, etc~\citep{yin2024survey}. Powered by cross-modal pretraining and instruction tuning, they now deliver strong or even superhuman performance. This rapid progress has encouraged a prevailing assumption: once perception is sufficiently strong, understanding will naturally emerge. In this story, the trajectory from perception to cognition appears to be a straightforward, even inevitable path toward general visual intelligence.

However, a deceptively simple failure, called the \textit{Watch Problem}, awakens the community from the comforting dream: current SoTA MLLMs consistently fail to generate a watch showing exactly six o'clock. Fig.~\ref{fig:watch} shows that MLLMs can name the watch, read digits, and list local cues such as the brand and the color of the strap, yet they fail to integrate those elements into a semantically correct representation of time.
This mismatch between detailed perception and correct interpretation revives a fundamental question: \textit{Can MLLMs truly understand what they can see?}

This question brings John Searle's Chinese Room argument~\citep{searle1982chinese} back into focus. Searle contends that perfectly rule-following symbol manipulation does not amount to genuine understanding. Extending this critique to the multi-modal domain, we propose the \textit{Visual Room} argument: \textit{an MLLM may precisely identify and even describe every visual detail, yet still miss the intended meaning}. In short, seeing is not understanding.

We introduce Visual Room 2.0, a hierarchical benchmark designed to explicitly disentangle perception from cognition and to assess their alignment. 
Given that both human perception and cognition are hierarchically organized in human information processing~\cite{rouw1997detecting,friston2010free}, we structure both perception and cognition into three progressive levels: low, middle (mid), and high, covering 17 representative tasks in total.
The perception track captures how models transform visual input into structured representations, progressing from low-level tasks {attribute recognition, sub-image detection}, through mid-level tasks {object detection, OCR}, to high-level tasks {scene description, scene understanding}.
In contrast, the cognition track assesses how models interpret and reason about meaning, advancing from low-level cognition {textual entailment, action recognition}, through mid-level cognition {emotion, sarcasm, humor understanding}, to high-level cognition {commonsense, causal, and social reasoning}.
The benchmark consists of 350 multi-modal samples, each paired with six progressive questions (for 2,100 in total), forming a vertical chain from perception to cognition (with one question for each level). 

We conduct a comprehensive evaluation on ten SoTA MLLMs (e.g., GPT-5, Qwen3-VL, Gemini 2.5, etc.), and compare their results across different tasks. The experimental results highlight three key findings: 
(1) MLLMs exhibit stronger perceptual competence than cognitive reasoning ability, with an average performance gap of 8.0\%.
(2) Cognition is not causally dependent on perception, as even under conditions of perfect perception, MLLMs fail 28.6\% of cognitive cases; and
(3) cognition scales with model size, but perception does not improve consistently with larger variants.
The main contributions of this work are:
\begin{itemize}
    \item We propose the \textit{Visual Room} argument, extending Searle's Chinese Room to the multi-modal domain.
    \item We introduce \textit{Visual Room 2.0}, a hierarchical benchmark spanning 17 perception and cognition tasks.
    \item Empirical findings revealing that current MLLMs remain perceptually competent but cognitively limited.
\end{itemize}

\section{Related Work}
The rapid progress of MLLMs has stimulated a surge of benchmarks aiming to quantify their visual understanding. Early efforts, such as MMBench~\cite{liu2024mmbench}, MMMU~\cite{yue2024mmmu}, and SuperCLUE-V\footnote{https://www.superclueai.com/homepage}, focus primarily on perceptual ability, namely evaluating how well MLLMs recognize, describe, or retrieve visual content. MMBench establishes a bilingual large-scale evaluation pipeline for visual instruction following. MMMU collects university-level questions to test cross-disciplinary knowledge reasoning, and SuperCLUE-V extends this to Chinese multimodal evaluation. Despite their scale and diversity, these benchmarks mostly assess surface-level perception rather than genuine understanding.

Subsequent benchmarks have begun to explore higher-order perception. II-Bench~\cite{2024II} evaluates models' grasp of abstract emotional and cultural meaning, while MVP-Bench~\cite{mvp}, PCA-Bench~\cite{pca}, and DeepEval~\cite{2024Can} propose more hierarchical evaluations that combine perception and reasoning tasks. However, these studies often blur the boundary between perception and cognition, implicitly assuming that advanced visual reasoning equates to genuine understanding. For example, PCA-Bench integrates perception, cognition, and action into a single chain without distinguishing their causal relations, and DeepEval examines deep semantics only at the perceptual level, without involving cognitive reasoning.

In contrast, Visual Room 2.0 explicitly separates perception from cognition and models each as a three-level hierarchy (low, middle, and high) across 17 tasks. This framework provides the first benchmark designed to measure whether MLLMs merely see or truly understand, as shown in Tab.~\ref{tab:compare}.
\begin{table}[t]
\centering
\caption{Comparison of Visual Room 2.0 with other benchmarks}\label{tab:compare}
\scalebox{0.75}{
\begin{tabular}{lccccc}
\toprule
\textbf{Dataset} & \textbf{Size} & \textbf{Perception} & \textbf{Cognition} & \textbf{Hierarchy} & \textbf{P–C Separation}\\
\midrule
MMBench~\cite{liu2024mmbench} &  3,000   &  \faCheckSquare  &  \faTimes  &  \faCheckSquare   &  \faTimes  \\
MMMU~\cite{yue2024mmmu} &  11.5K  &  \faCheckSquare   &  \faTimes  &  \faTimes   &  \faTimes  \\
\midrule
II-Bench\cite{2024II} &  1,222   &  \faCheckSquare  &  \faCheckSquare &  \faTimes  &  \faTimes    \\
MVP-Bench\cite{mvp} &  1,060   &  \faCheckSquare  &  \faCheckSquare &  \faCheckSquare  &  \faTimes     \\
PCA-Bench\cite{pca} &  7,510   &  \faCheckSquare  &  \faCheckSquare &  \faTimes  &  \faCheckSquare  \\
DeepEval\cite{2024Can}  & 1,001 &  \faCheckSquare  &  \faCheckSquare &  \faCheckSquare  &  \faTimes    \\\midrule
\textbf{Visual Room} &  2,100  &  \faCheckSquare  & \faCheckSquare &  \faCheckSquare  &  \faCheckSquare   \\
\bottomrule
\end{tabular}}
\end{table}

\section{The Visual Room 2.0 Benchmark}
\subsection{Visual Room Argument}
Inspired by Searle's Chinese Room, we extend the classic thought experiment into the multi-modal domain and propose the \textit{Visual Room} argument (as shown in Fig.~\ref{fig:visualroom} in App.~A): \textit{Imagine an operator locked in a sealed room, mechanically following manuals to describe every visual detail, e.g., objects, colors, and expressions, without any true grasp of their meaning. To an observer, the system seems to ``see'' and ``understand'', yet internally it only manipulates symbols according to predefined rules.}

This argument motivates the design of Visual Room 2.0 Benchmark, which transforms the philosophical paradox into a measurable evaluation framework by hierarchically separating perception and cognition. Through structured, multi-level tasks, the benchmark allows us to empirically locate where perception succeeds but understanding fails.

\subsection{Benchmark Construction}
\textbf{Data acquisition.} To ensure that the dataset captures both perceptual and cognitive characteristics, we selected \textit{Reddit} as our primary data source.
Unlike other social platforms such as Facebook, Instagram, or Weibo, Reddit organizes its content into topical subreddits, which allows precise sampling within well-defined discourse contexts.
We focus on four cognition-rich subreddits: i.e., r/sad, r/happy, r/humor, and r/sarcasm, which naturally span a spectrum of emotional and interpretive phenomena from basic affect recognition to complex sarcasm understanding.
We define three selection criteria:
(1) each post contains both an image and accompanying text to ensure multi-modal completeness;
(2) the author explicitly expresses an emotion or attitude in the title or comments (e.g., a ``sarcasm'' tag) for reliable labeling;
(3) each post receives at least three comments or thirty upvotes as a weak supervision signal of social consensus.
In total, we collect 2,000 multi-modal samples.

\textbf{Data filtering.} We adopt a three-stage filtering pipeline: (1) we remove images where more than 40\% of the area is occupied by text; (2) we discard samples containing only a single object or a plain background; (3) we manually verify that the image and its accompanying text are semantically coherent.

After filtering, we retain 350 high-quality multi-modal samples for Visual Room 2.0.

\begin{figure}[t!]
    \centering
    \includegraphics[width=0.45\textwidth]{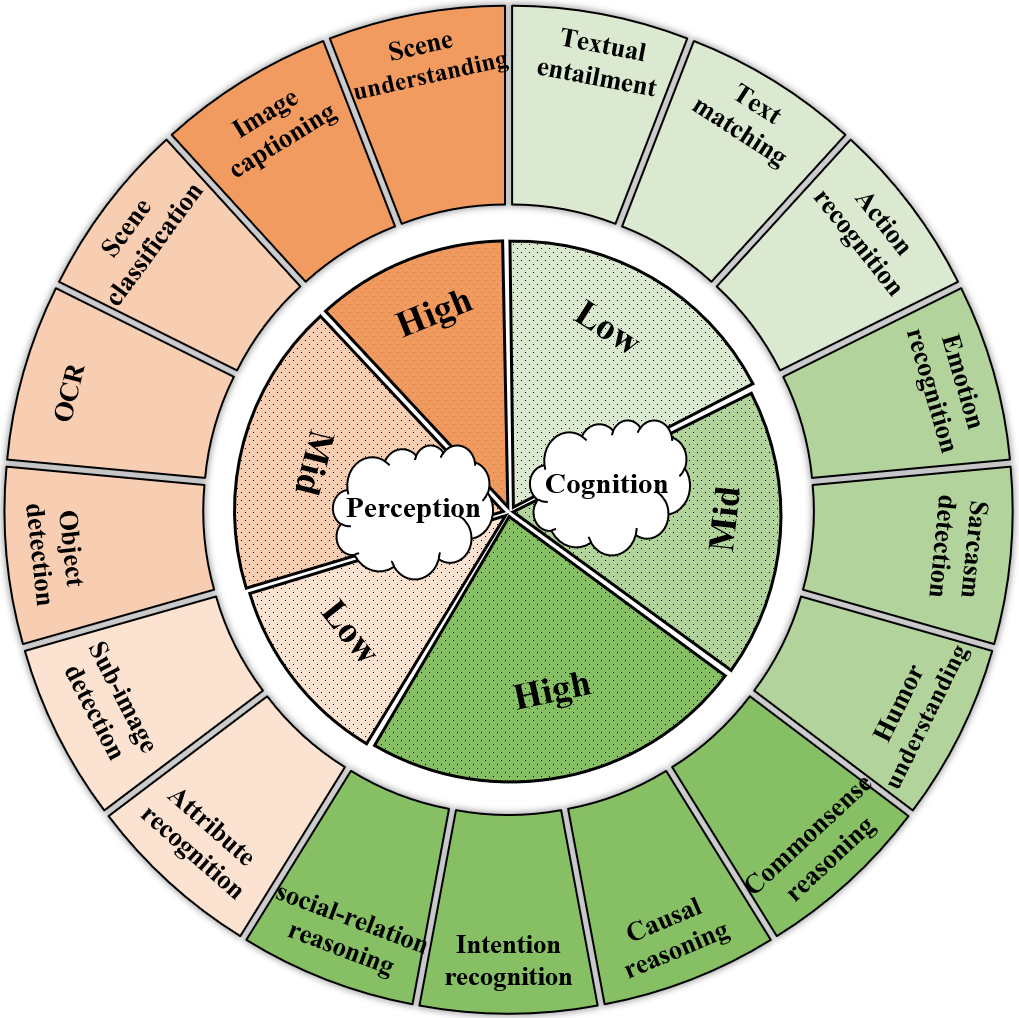}
    \caption{Definition of high-, mid- and low-level perception and cognition.}
    \label{fig:distribution1}
\end{figure}

\textbf{Task definition.} We organize the benchmark according to human information-processing theory by dividing both perception and cognition into three hierarchical levels: low, middle, and high.
The \textbf{perception} hierarchy progresses from direct feature sensing to holistic scene understanding: the low-level stage focuses on basic visual attributes such as color, shape, and texture (e.g., \textit{attribute recognition, sub-image detection}); the mid-level stage emphasizes object- and structure-centered recognition (e.g., \textit{object detection}, \textit{optical character recognition}, and \textit{scene classification}); and the high-level stage integrates objects, relations, and contextual cues for global comprehension (e.g., \textit{image captioning} and \textit{scene understanding}), as shown in Fig.~\ref{fig:distribution1}.

Similarly, the \textbf{cognition} hierarchy reflects ascending levels of semantic and reasoning complexity: the low-level stage targets basic symbolic reasoning and semantic matching (e.g., \textit{textual entailment}, \textit{text matching}, and \textit{action recognition}); the mid-level stage involves multi-modal semantic fusion and emotional inference (e.g., \textit{emotion recognition}, \textit{sarcasm detection}, and \textit{humor understanding}); and the high-level stage engages human-like reasoning over common sense, causality, intention, and social relationships (e.g., \textit{commonsense reasoning}, \textit{causal reasoning}, \textit{intention recognition}, and \textit{social-relation reasoning}).

\textbf{Question design.} Each sample contains six questions aligned with the six perception–cognition levels.
Five multiple-choice questions correspond to low- and mid-level perception, and low-, mid-, and high-level cognition.
The remaining short-answer question represents high-level perception, requiring a fine-grained scene description that captures key visual elements and their spatial relations.
Scene descriptions are generated via a ``Claude-4 + human verification'' pipeline, following strict guidelines to ensure objectivity, coverage, and clarity (third-person perspective, no emotional terms, within 50 words).

\textbf{Quality control.} Three annotators conduct the main labeling. Before formal annotation, they complete 20 pilot samples targeting 90\% agreement.
After full annotation, we compute Cohen's $\kappa$ = 0.73, indicating strong inter-annotator reliability.
All questions are further reviewed by a fourth verifier to standardize difficulty and linguistic style.

\begin{figure}[t!]
    \centering
    \includegraphics[width=0.51\textwidth]{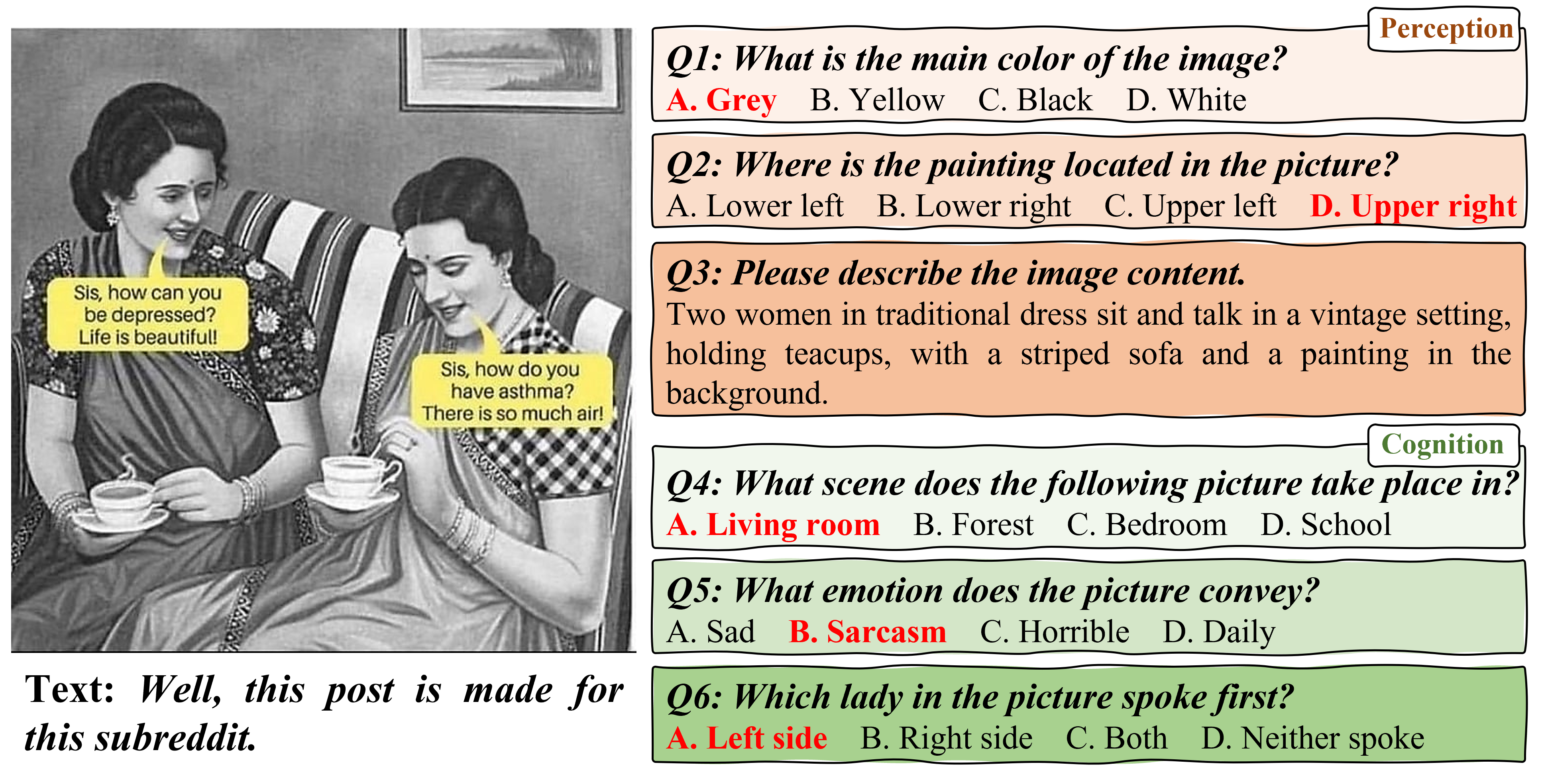}
    \caption{An example of visual room 2.0 benchmark.}
    \label{fig:example}
\end{figure}

Finally, Visual Room 2.0 contains 350 multi-modal samples × 6 questions = 2,100 questions, systematically spanning perception and cognition tasks, as illustrated in Fig.~\ref{fig:example} and Tab.~\ref{tab:statistics}.

\subsection{Benchmark Analysis}
\subsubsection{\textbf{Task distribution analysis.}}
The Visual Room 2.0 benchmark comprises 1,050 perception and 1,050 cognition questions, systematically covering 17 representative tasks. 

In the \textbf{perceptual hierarchy}, the majority of samples center on \textit{attribute recognition} (31.2\%) and \textit{object detection} (19.2\%), constituting the most frequent perceptive types. Meanwhile, \textit{image captioning} (18.7\%) and \textit{scene understanding} (14.7\%) represent higher-order perceptual integration, requiring multi-object reasoning and compositional semantics. The smaller proportions of \textit{scene classification} (7.6\%), \textit{OCR} (6.5\%), and \textit{sub-image detection} (2.1\%) indicate that highly localized or text-dense perception tasks are less emphasized, aligning with our focus on holistic visual comprehension.  

In the \textbf{cognitive hierarchy}, the task design prioritizes inferential and affective reasoning, which are central to human-like understanding. \textit{Intention recognition} (15.9\%) and \textit{emotion recognition} (15.1\%) constitute the most frequent cognitive types, underscoring the role of social and emotional inference in bridging perception and cognition. \textit{Text matching} and \textit{sarcasm detection} (13.9\% and 10.3\%) capture mid-level semantic integration, while \textit{humor understanding}, \textit{causal reasoning}, and \textit{commonsense reasoning} (7--8\%) extend to abstract pragmatic inference. 

Overall, this stratified task allocation ensures balanced coverage across perceptual granularity and cognitive abstraction, forming a coherent testbed for analyzing how MLLMs transition from \textit{seeing} to \textit{understanding}.


\begin{table}[t!]
\centering
\caption{Statistics Summary}
\small
\renewcommand{\arraystretch}{1.1} 
\begin{minipage}[t]{0.0\textwidth}
    \centering
    \begin{tabular}{p{2.5cm} r}
    \toprule
    \textbf{Item} & \textbf{Num} \\
    \midrule
    Image & 350 \\
    Question & 2100 \\
    \toprule 
    \textbf{Item} & \textbf{Avg} \\
    \midrule
    Text Length	& 14.42 \\
    Question length & 13.76 \\
    Description length & 23.66 \\
    \toprule
    \textbf{Perception Task} & \textbf{Num} \\
    \midrule
    Attribute recognition & 328 \\
    Sub-image detection & 22 \\
    Object detection & 202 \\
    OCR & 68 \\
    Scene classification & 80 \\
    Image captioning &  196\\
    Scene understanding & 154 \\
    \bottomrule
    \end{tabular}
\end{minipage}
\hfill 
\begin{minipage}[t]{0.25\textwidth} 
    \centering 
    \begin{tabular}{p{2.9cm} r}
    \toprule
    \textbf{Cognition Task} & \textbf{Num} \\
    \midrule
    Textual entailment & 103 \\
    Text matching & 146 \\
    Action recognition & 101 \\
    Emotion recognition & 159 \\
    Sarcasm detection & 108 \\
    Humor understanding & 83 \\
    Commonsense reasoning & 65 \\
    Causal reasoning & 72 \\
    Intention recognition & 167 \\
    Social-relation reasoning & 46 \\
    \toprule
    \textbf{Emotion Type} & \textbf{Num} \\
    \midrule
    Happy & 99 \\
    Sad & 60 \\
    Humor & 83 \\
    Sarcasm & 108 \\
    \bottomrule
    \end{tabular}
\end{minipage}

\label{tab:statistics} 
\end{table}

\subsubsection{\textbf{Hierarchical correlation analysis.}}
\begin{figure}[t!]
    \centering
    \includegraphics[width=0.5\textwidth]{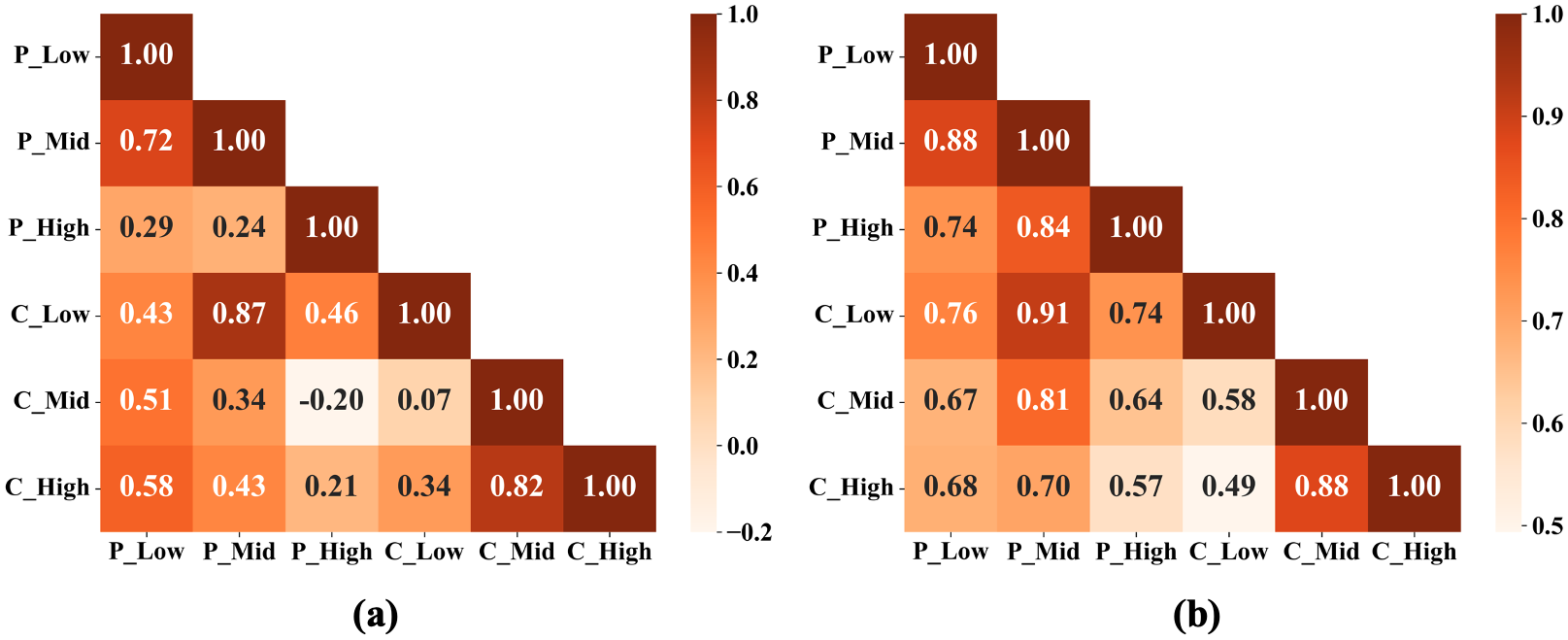}
    \caption{Fig.~(a) represents Spearman Correlation between different levels, and (b) represents Pearson Correlation.}
    \label{fig:rador1}
\end{figure}

To verify whether the proposed three-level division (\textit{low}, \textit{mid}, \textit{high}) within perception and cognition tasks reflects a valid hierarchical structure, we compute both Pearson and Spearman correlation coefficients across all levels (see Tab.~\ref{tab:pcorrelation} and Fig.~\ref{fig:rador1}). Pearson correlations measure linear consistency of model scores, whereas Spearman correlations evaluate rank stability to ensure robustness.
For perception, the correlations exhibit a clear decreasing pattern across levels: \textbf{Low–Mid} ($r_{p}=0.883,\,p<0.01$) $>$ \textbf{Mid–High} ($r_{p}=0.841,\,p<0.01$) $>$ \textbf{Low–High} ($r_{p}=0.737,\,p<0.05$).  
This monotonic decline indicates a well-defined difficulty gradient—model performance decreases progressively with increasing perceptual complexity.  
However, the corresponding Spearman correlations drop markedly at higher levels (e.g., $\rho_{Mid–High}=0.237$), suggesting that high-level perception reorders model ranking and thus recruits partially distinct perceptual abilities beyond linear scaling of lower-level skills.

Cognition reveals a different structure.  
Pearson results show moderate Low–Mid correlation ($r_{p}=0.583,\,p=0.08$) but strong Mid–High correlation ($r_{p}=0.879,\,p<0.01$), while Spearman correlations mirror this two-phase pattern ($\rho_{Low–Mid}=0.073$, $\rho_{Mid–High}=0.818,\,p<0.01$).  
These results indicate that low-level cognitive tasks rely on relatively simple pattern-matching or literal reasoning, while mid- and high-level tasks require more complex inferential processes such as causal and social reasoning, leading to more consistent model performance across advanced levels.

Overall, the perception hierarchy demonstrates a \textit{continuous gradient}, whereas the cognition hierarchy exhibits a \textit{stage-wise clustering}.  
Together, these trends empirically validate the rationality of our three-tier design and highlight a fundamental structural asymmetry between perception and cognition—linear progression versus non-linear abstraction, which aligns with the core intuition of the \textit{Visual Room} argument.
\begin{table}[t!]
\centering
\caption{Correlation analysis across hierarchical levels.}\label{tab:pcorrelation}
\scalebox{0.78}{
\begin{tabular}{lccccc}
\toprule
\multirow{2}{*}{\textbf{Level Pair}} & \multicolumn{2}{c}{\textbf{Perception}} & & \multicolumn{2}{c}{\textbf{Cognition}} \\
\cline{2-3} \cline{5-6}
 & \textbf{Pearson $r$ / $p$} & \textbf{Spearman $r$ / $p$} & & \textbf{Pearson $r$ / $p$} & \textbf{Spearman $r$ / $p$} \\
\midrule
Low--Mid  & \textbf{0.883 / 0.0007} & \textbf{0.720 / 0.0190} & & 0.583 / 0.0766 & 0.073 / 0.8408 \\
Mid--High & 0.841 / 0.0023 & 0.237 / 0.5096 & & \textbf{0.879 / 0.0008} & \textbf{0.818 / 0.0038} \\
Low--High & 0.737 / 0.0151 & 0.286 / 0.4236 & & 0.494 / 0.1470 & 0.335 / 0.3435 \\
\bottomrule
\end{tabular}
}
\end{table}

\begin{table}
\centering
\caption{Comparison of 10 MLLMs on Visual Room 2.0. \underline{Underline} indicates the best, \textit{italic} indicates the second, and \textbf{Bold} indicates the Average.}
\scalebox{0.93}{
\begin{tabular}{lcccccc}
\toprule
\multirow{2}{*}{\textbf{Models}} & \multicolumn{3}{c}{\textbf{Perception}} & \multicolumn{3}{c}{\textbf{Cognition}}\\
\cmidrule(r){2-4} \cmidrule(r){5-7}
& \textbf{Low} & \textbf{Mid} & \textbf{High} & \textbf{Low} & \textbf{Mid} & \textbf{High} \\
\midrule
Qwen 2.5 VL-7B & 
86.50&	82.00& \textit{\textbf{74.25}}&	82.00&	61.00&	53.75\\
Deepseek VL-7B & 80.25&	67.75&	46.75&	76.00&	51.75&	48.75\\
\midrule
GPT-5 & \textit{\textbf{87.00}} &	84.75& 69.00& 82.00&	\underline{\textbf{79.00}}&	\textit{\textbf{67.75}} \\
GPT-o1 & 86.75&	85.50&	65.75&	82.25&	73.82&	60.75 \\
GPT-4V & 84.25&	82.00&	73.25&	82.25&	71.25&	60.67 \\
GLM-4V-Plus & 86.00 &	85.00&	66.50&	84.25&	72.75&	65.75\\
Qwen-VL-Max &
84.75& 	83.25& 73.75& 	82.50& 	71.77& 	61.32 \\
Gemini 2.5 Pro & \underline{\textbf{88.50}} & \underline{\textbf{89.50}} &  \underline{\textbf{77.00}} & \underline{\textbf{86.50}} & 72.25 & \underline{\textbf{68.20}} \\
Gemini 2.5 Flash & 
84.75&	\textit{\textbf{87.25}} & 72.25 &	\textit{\textbf{85.50}}&	71.45&	57.28 \\
Doubao-vision Pro &
84.60&	81.90&	71.79&	79.26&	\textit{\textbf{73.97}}&	67.27\\
\toprule
\textbf{Average} & \textbf{85.34} & \textbf{82.89} & \textbf{69.03} & \textbf{82.25} & \textbf{69.90} & \textbf{61.15} \\
\textbf{$\bigtriangleup \left |M/H-L\right |$   } & - & \textbf{2.45} & \textbf{16.31} & -
& \textbf{12.35} & \textbf{21.10} \\
\bottomrule
\end{tabular}}
\label{tab:main_results}
\end{table}

\section{Experiments and Analysis}
\subsection{Experimental Settings}
We conduct evaluation experiments on Visual Room 2.0 over 10 SoTA LLMs. They are: (1) \textbf{\underline{GPT-5}}, (2) \textbf{\underline{GPT-o1}}, (3) \textbf{\underline{GPT-4V}}, (4) \textbf{\underline{GLM-4V-Plus}}, (5) \textbf{\underline{Qwen-VL-Max}}, (6) \textbf{\underline{Gemini 2.5 Flash}}, \\(7) \textbf{\underline{Gemini 2.5 Pro}}, (8) \textbf{\underline{Doubao-vision Pro}} are eight big models. In contrast, (9) \textbf{\underline{Qwen 2.5 VL-7B}} and (10) \textbf{\underline{DeepSeek VL-7B}} are two SoTA small models.

\subsection{Evaluation and Metrics}
All MLLMs are evaluated under zero-shot standard I/O prompting. Each sample is tested three times, and the average score is reported.
For multiple-choice questions, we use \textit{Accuracy} as the metric.
For short-answer questions, we compute semantic similarity between LLMs' output and human annotation using a hybrid similarity metric (details please see App.~C).

\subsection{Main Results}
Tab.~\ref{tab:main_results} reports independent evaluations of ten MLLMs on Visual Room 2.0.
Both perception and cognition tracks are assessed independently at each level (Low, Mid, High) with no shared samples or contextual prompts.

Overall, all MLLMs perform significantly better on perception tasks than on cognition tasks, with mean accuracies of \textbf{79.09\%} and \textbf{71.10\%}, respectively. 
This indicates that current MLLMs can effectively recognize visual content and scene semantics but still struggle with higher-order understanding such as emotion, intention, and causal reasoning, supporting the hypothesis that \textbf{seeing $\neq$ understanding}.

Across difficulty levels, performance decreases consistently as task complexity increases. 
In perception, the average accuracy drops by \textbf{16.3\%} from low- to high-level tasks, while cognition shows an even larger decline of \textbf{21.1\%}. 
This gradient confirms that cognitive tasks impose greater challenges than perceptual ones, especially for high-level reasoning tasks (e.g., causal and social-relation inference).

Regarding model scale, clear differences emerge. 
\textbf{Smaller models} (e.g., \textit{Qwen 2.5 VL-7B}) achieve perception accuracy comparable to large models such as \textit{Gemini 2.5 Pro} and \textit{GPT-5} 
(e.g., \textbf{86.5\%} and \textbf{74.25\%} on low and high levels, respectively). 
However, their performance drops sharply on cognitive tasks (\textbf{61.0\%} on mid-level and \textbf{53.75\%} on high-level), revealing a pronounced gap between recognition and reasoning. 
In summary, model scaling has a limited effect on perceptual performance, as even smaller models achieve results comparable to larger ones.
However, cognition benefits substantially from scale, with larger models demonstrating markedly higher accuracy.

\subsection{Cognitive Performance under Correct Perception}
Tab.~\ref{tab:cog} presents the cognitive results of ten MLLMs evaluated under the condition of perfect perception, where only samples with entirely correct perceptual responses are considered. 
This experiment tests whether accurate perception causally enhances cognitive reasoning. 
The results show no consistent improvement: the average cognitive accuracy changes only slightly by \textbf{+0.09\%} (Low), \textbf{+1.31\%} (Mid), and \textbf{-0.37\%} (High). 
Even with flawless perception, most models fail to achieve higher understanding, indicating that correct visual encoding alone does not necessarily translate into better reasoning. 
Several models, such as \textit{Doubao-vision Pro} and \textit{GPT-4V}, even decline slightly, suggesting that current architectures struggle to leverage perceptual information for higher-level inference.

Large models (\textit{GPT-o1},\textit{Gemini 2.5 Pro},  \textit{Gemini 2.5 Flash}) show minor gains in high-level cognition (\textbf{+0.30\%}–\textbf{+0.63\%}),
whereas smaller ones like \textit{Qwen 2.5 VL-7B} and \textit{DeepSeek VL-7B} remain stagnant or progress. 
On average, MLLMs still fail in about \textbf{28.6\%} of cognitive cases in average, even when perception is perfect.

\begin{table}[htbp]
\centering
\caption{\textit{Cognitive} performance under correct perception.}
\label{tab:cog}
\resizebox{0.45\textwidth}{!}{ 
\begin{tabular}{lccc}
\toprule
\textbf{Models} & \textbf{Low} & \textbf{Mid} & \textbf{High} \\
\midrule
Qwen 2.5 VL-7B & 83.12 \textcolor{green}{(1.12 $\uparrow$)} & 61.01 \textcolor{green}{(0.01 $\uparrow$)} & 53.11 \textcolor{red}{(0.64 $\downarrow$)} \\
DeepSeek VL-7B & 77.71 \textcolor{green}{(1.71 $\uparrow$)} & 53.60 \textcolor{green}{(1.85 $\uparrow$)} & 52.32 \textcolor{green}{(3.57 $\uparrow$)} \\
\midrule
GPT-5 & 82.96 \textcolor{green}{(0.96 $\uparrow$)} & 79.47 \textcolor{green}{(0.47 $\uparrow$)} & 67.15 \textcolor{red}{(0.60 $\downarrow$)} \\
GPT-o1 & 83.15 \textcolor{green}{(0.90 $\uparrow$)} & 76.77 \textcolor{green}{(2.95 $\uparrow$)} & 61.38 \textcolor{green}{(0.63 $\uparrow$)} \\
GPT-4V & 83.63 \textcolor{green}{(1.38 $\uparrow$)} & 70.60 \textcolor{red}{(0.65 $\downarrow$)} & 59.31 \textcolor{red}{(1.36 $\downarrow$)} \\
GLM-4V-Plus & 83.17 \textcolor{red}{(1.08 $\downarrow$)} & 75.35 \textcolor{green}{(2.60 $\uparrow$)} & 64.66 \textcolor{red}{(1.09 $\downarrow$)} \\
Qwen-VL-Max & 83.34 \textcolor{green}{(0.84 $\uparrow$)} & 74.17 \textcolor{green}{(2.40 $\uparrow$)} & 61.00 \textcolor{red}{(0.32 $\downarrow$)} \\
Gemini 2.5 Pro & 88.00 \textcolor{green}{(1.50 $\downarrow$)} & 73.59 \textcolor{green}{(1.34 $\downarrow$)} & 68.50 \textcolor{green}{(0.30 $\uparrow$)} \\
Gemini 2.5 Flash & 83.28 \textcolor{red}{(0.22 $\downarrow$)} & 71.69 \textcolor{green}{(0.24 $\uparrow$)} & 57.82 \textcolor{green}{(0.54 $\uparrow$)} \\
Doubao-vision Pro & 75.06 \textcolor{red}{(4.20 $\downarrow$)} & 75.83 \textcolor{green}{(1.86 $\uparrow$)} & 62.50 \textcolor{red}{(4.77 $\downarrow$)} \\
\toprule
\textbf{Average} & \textbf{82.34 \textcolor{green}{(0.09 $\uparrow$)}} & \textbf{71.21 \textcolor{green}{(1.31 $\uparrow$)}} & \textbf{60.78 \textcolor{red}{(0.37 $\downarrow$)}} \\
\bottomrule
\end{tabular}}
\end{table}

\subsection{Difficulty Gradient Analysis}
Tab.~\ref{tab:diff} presents the inter-level performance gaps across perception and cognition. 
A clear asymmetry emerges between the two domains. 
For \textit{perception}, the gap from Low to Mid is small (\textbf{2.45 pts}), whereas the gap from Mid to High is large (\textbf{13.86 pts}); this indicates that difficulty concentrates at the transition to holistic, scene-level understanding. By contrast, for \textit{cognition}, the largest drop occurs from Low to Mid (\textbf{12.35 pts}), while the additional drop from Mid to High is more modest (\textbf{8.75 pts}), despite a sizable High–Low difference of \textbf{21.10} points overall; this suggests that cognitive demands escalate sharply once tasks require affective/pragmatic integration, after which performance degrades more gradually rather than collapsing further.

\begin{table}[htbp]
\centering
\caption{Inter-level performance differences across perception and cognition.}
\label{tab:diff}
\begin{tabular}{lcc}
\toprule
 & \textbf{Perception} & \textbf{Cognition}\\
\midrule
Low & 85.34 & 82.25 \\
Mid & 82.89 & 69.9 \\

\rowcolor{gray!40} \textbf{$\left |Mid-Low\right |$   }& 2.45 & 12.35 \\
 \toprule
Mid & 82.89 & 69.9 \\
High & 69.03 & 61.15 \\
\rowcolor{gray!40}
 $\left |High-Mid\right |$ & 13.86 & 8.75 \\
 \toprule
 Low & 85.34 & 82.25 \\
 High & 69.03 & 61.15 \\
\rowcolor{gray!40}
 $\left |High-Low\right |$ & 16.31	&21.1\\
\bottomrule
\end{tabular}
\end{table}

\subsection{Effect of Model Scaling on Perception and Cognition}
\begin{figure}[htbp]
    \centering
    \includegraphics[width=1\linewidth]{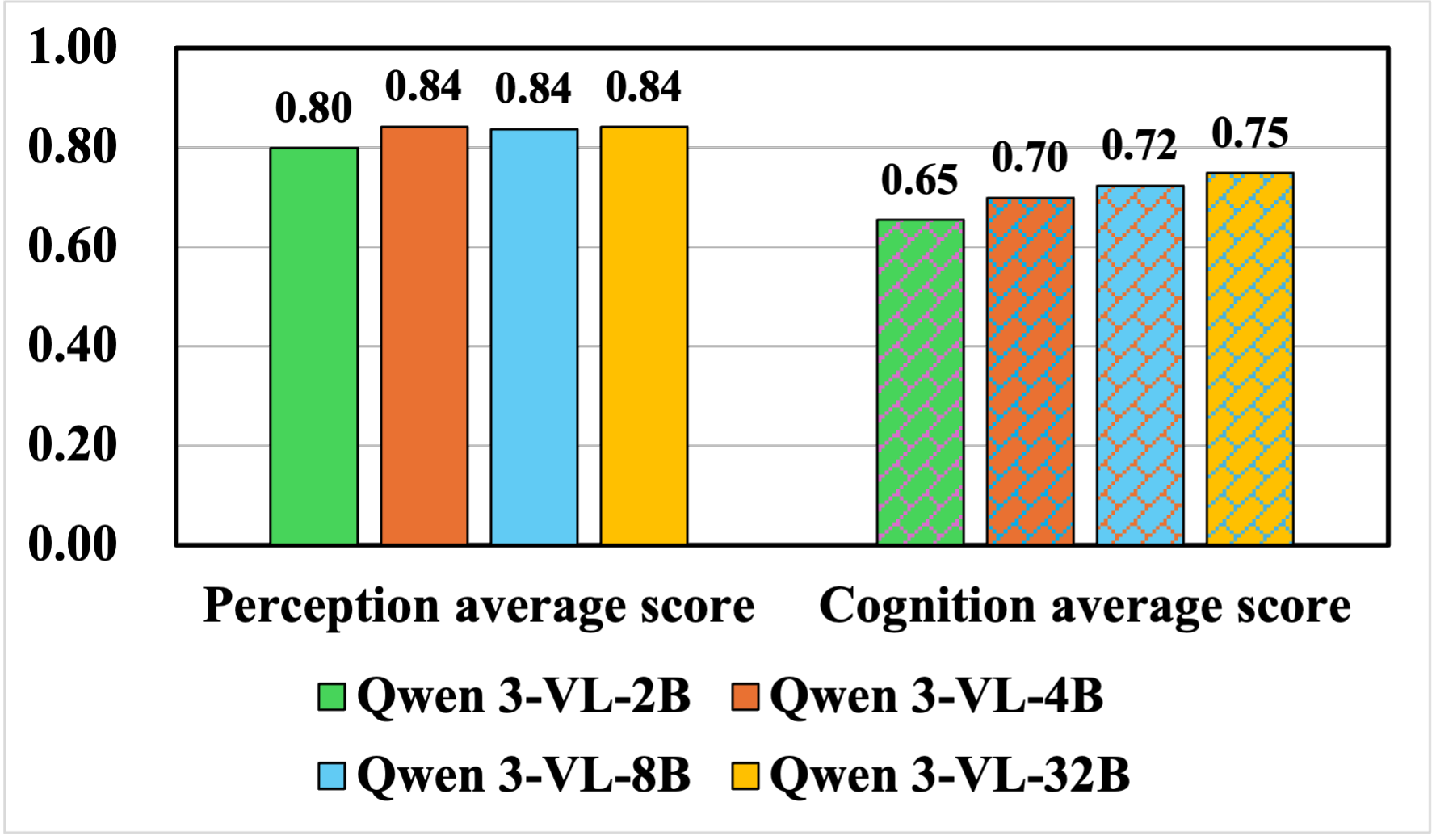}
    \caption{Effect of model scaling on perception and cognition.}
    \label{fig:scale}
\end{figure}

Fig.~\ref{fig:scale} illustrates the influence of model size on perceptual and cognitive performance across the Qwen3-VL series (2B/4B/8B/32B). 
A distinct divergence emerges between the two dimensions. \textit{Perception} performance remains nearly saturated, increasing only slightly from 0.80 (2B) to 0.84 (32B), suggesting that low-level visual recognition benefits minimally from larger parameter counts. In contrast, \textit{cognition} improves consistently from 0.65 to 0.75 as model size grows, reflecting stronger abstraction, cognition, and contextual integration capabilities. This widening gap indicates that scaling primarily enhances higher-order reasoning rather than perceptual fidelity. The results further imply that perceptual modules in current MLLMs have reached a structural bottleneck, whereas cognitive understanding continues to rely on broader representational capacity and deeper semantic alignment.

\subsection{Perception–Cognition Relationship Analysis}

\begin{figure}[htbp]
    \centering
    \includegraphics[width=0.9\linewidth]{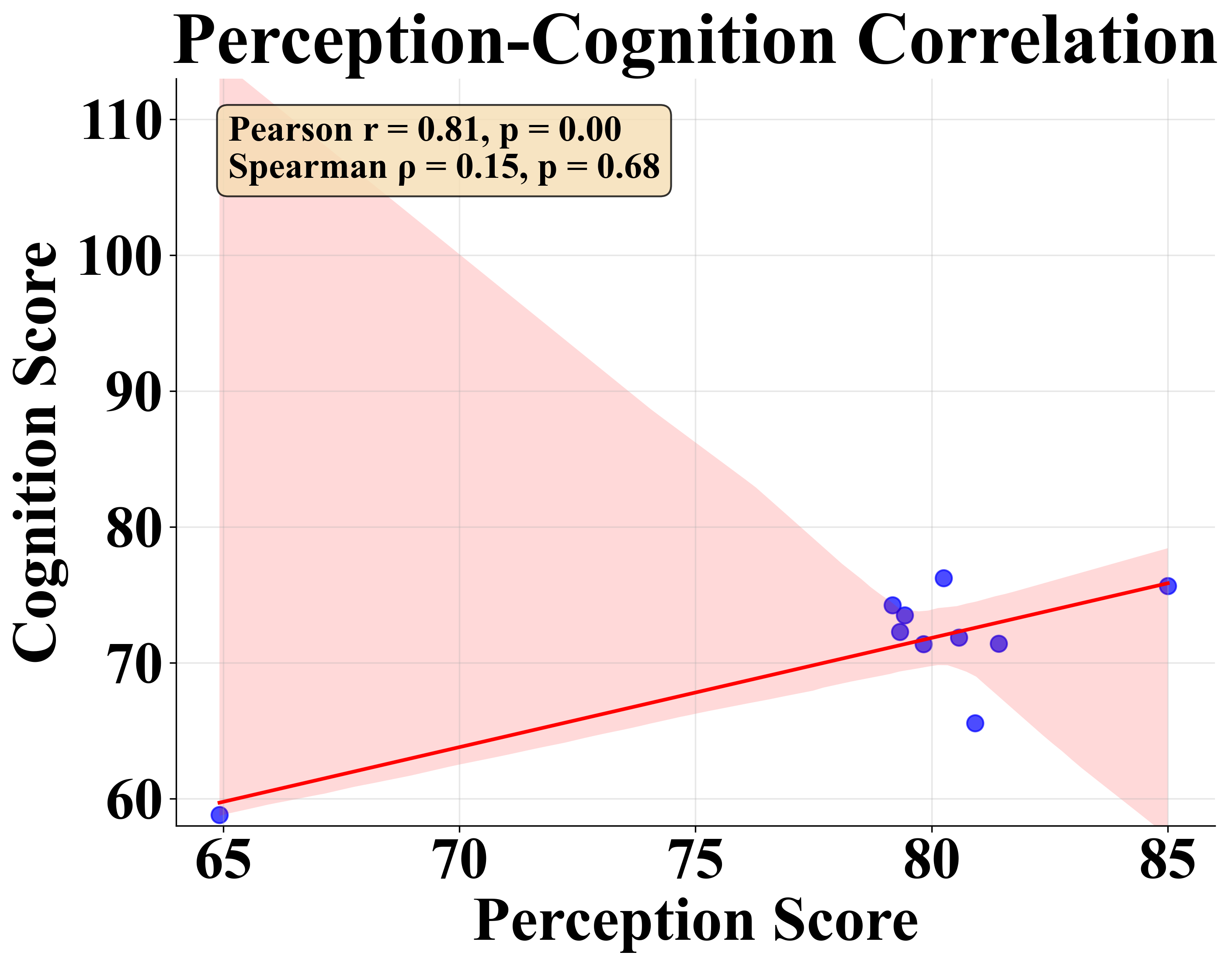}
    \caption{Perception cognition relationship.}
    \label{fig:pc_corr}
\end{figure}

To further examine whether perceptual competence causally contributes to cognitive reasoning, we conducted a correlation analysis between the average perception and cognition scores across all ten MLLMs, as shown in Fig.~\ref{fig:pc_corr}. 
The results reveal a strong linear association (\textbf{Pearson} $r = 0.81$, $p < 0.01$) but a weak and non-significant monotonic correlation (\textbf{Spearman} $\rho = 0.15$, $p = 0.68$). 
This discrepancy indicates that while LLMs with higher overall capacity tend to perform better in both perception and cognition, the ranking consistency across the two domains is unstable. 
In other words, perceptual accuracy does not necessarily entail stronger reasoning ability. 
The correlation pattern suggests that perception and cognition are not causally linked but instead co-vary as functions of general representational capacity—supporting the core assumption of the \textit{``Seeing $\neq$ Understanding''} hypothesis. 

\subsection{Does explicit reasoning help?}
\begin{figure}[htbp]
    \centering
    \includegraphics[width=1\linewidth]{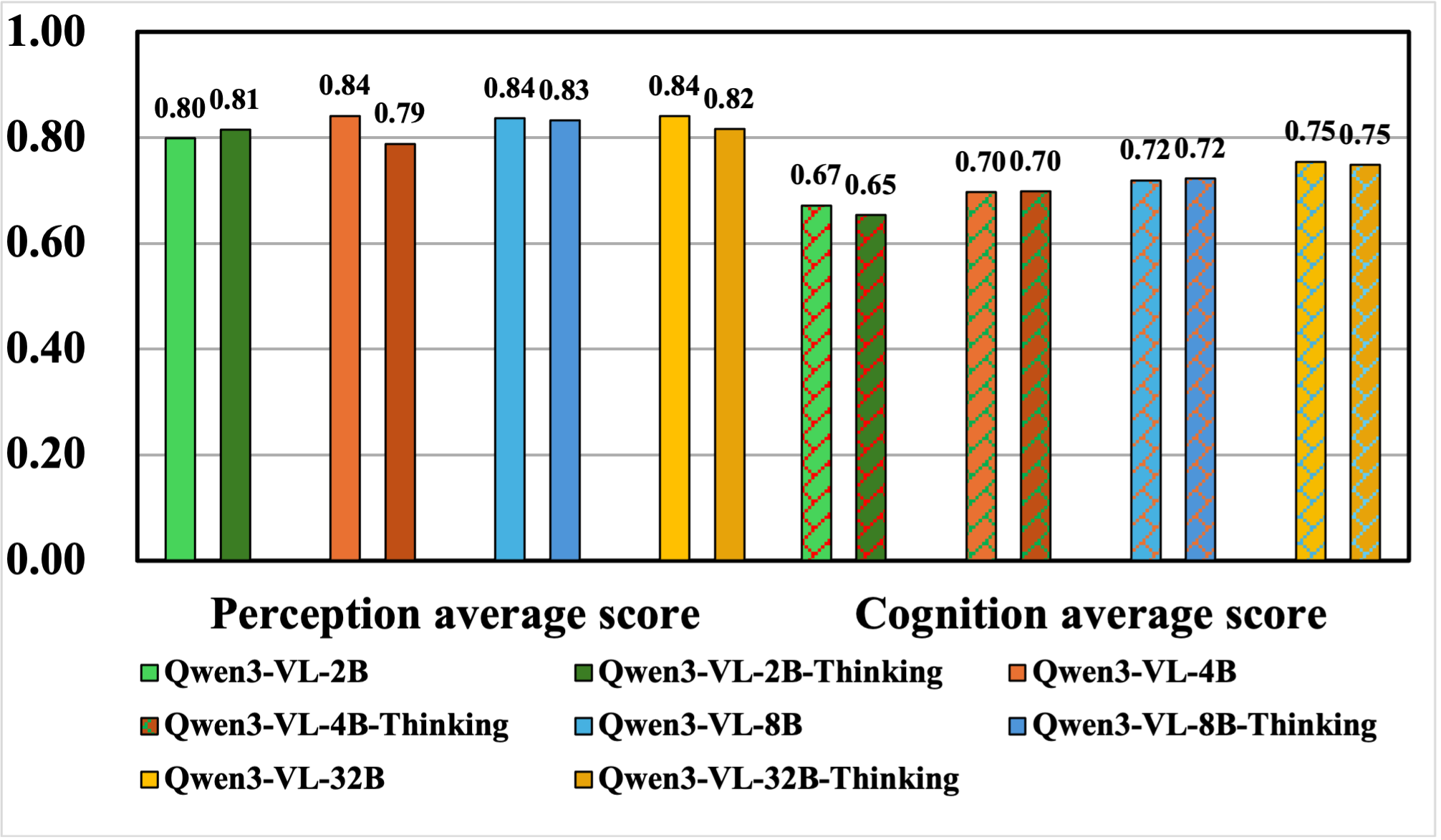}
    \caption{Comparison of base models and reasoning models.}
    \label{fig:reason}
\end{figure}

We compare Qwen3-VL base models (2B/4B/8B/32B) with their \textit{Thinking} variants (i.e., long chain-of-thought) under identical evaluation. Fig.~\ref{fig:reason} shows no consistent benefit from explicit reasoning. 
For \textbf{perception}, performance changes are inconsistent: $+1.6\%$ (2B), $-5.3\%$ (4B), $-0.4\%$ (8B), and $-2.4\%$ (32B), averaging \textbf{$-1.6\%$}. 
For \textbf{cognition}, changes remain marginal: $-1.8\%$, $+0.2\%$, $+0.3\%$, and $-0.5\%$ across the same models, averaging \textbf{$-0.45\%$}. 
Although moderate improvements appear in the 4B and 8B models, they are offset by declines at smaller (2B) and larger (32B) scales, indicating that long CoT reasoning neither enhances visual–semantic grounding nor consistently improves higher-order inference. 
Overall, explicit reasoning offers limited benefit and may even introduce instability in perceptual accuracy, suggesting that robust cognition in MLLMs requires deeper structural integration rather than procedural reasoning prompts.

\subsection{Case Study}
\begin{figure}[htbp]
    \centering
\includegraphics[width=1\linewidth]{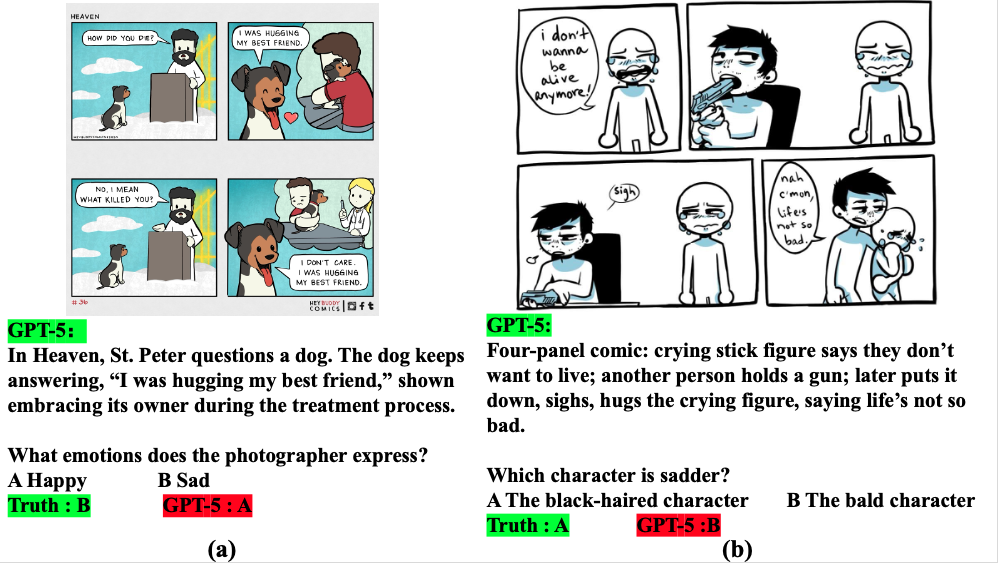}
    \caption{Case Study}
    \label{case}
\end{figure}

We conduct a case study, as illustrated in Fig.~\ref{case}, to further examine the discrepancy between perceptual accuracy and cognitive reasoning. We selected instances from the GPT-5 model where perceptual processing was accurate but cognitive reasoning failed. In the Fig.\ref{case}(a), the model accurately identified the dog's cheerful expression and the joyful tone of the dialogue; however, it failed to infer the underlying sadness implied by the contextual events. Similarly, in the Fig.\ref{case}(b), the model correctly recognized that the bald character shed tears in the first and third frames, while the black-haired character, though visibly in pain, did not. Nonetheless, the model was unable to comprehend the deeper emotional implication behind the black-haired character’s suicide attempt. 
In summary, MLLM may precisely identify and even describe every visual detail (e.g., \textit{facial expressions}, \textit{text} and \textit{scenes}) , but it is difficult to combine them to understand the complex implied meanings behind them.

\section{Conclusion}
In this paper, we extend Searle's Chinese Room argument to the multi-modal domain and propose the Visual Room argument, which distinguishes mechanical perception from genuine understanding in MLLMs.
To operationalize this idea, we introduce Visual Room 2.0, a hierarchical benchmark designed to disentangle perception from cognition across 17 representative tasks.
Through systematic evaluation of ten SoTA MLLMs, we reveal a persistent perception–cognition gap: strong perceptual accuracy does not guarantee semantic or pragmatic understanding, and even under perfect perception, models still fail in a substantial portion of cognitive tasks.
This suggests that perception and cognition are functionally independent rather than causally linked.
This work operationalizes ``Seeing $\ne$ Understanding'' as a testable hypothesis, offering a new paradigm from perceptual processing to cognitive reasoning in MLLMs.

\textbf{Limitations.} Despite its contributions, this study has two main limitations.
First, although the benchmark contains 2,100 annotated questions across 350 multimodal samples, it still does not cover the full spectrum of perceptual and cognitive tasks.
Second, the current evaluation relies mainly on multiple-choice and short-answer questions for static image understanding, without incorporating other modalities such as audio or video.
Future work will extend Visual Room 2.0 to multi-modal, dynamic, and interactive contexts for a more comprehensive assessment.

\bibliographystyle{ACM-Reference-Format}
\bibliography{references}
\clearpage

\appendix
\section{Visual Room}
\begin{figure}
    \centering
    \includegraphics[width=1\linewidth]{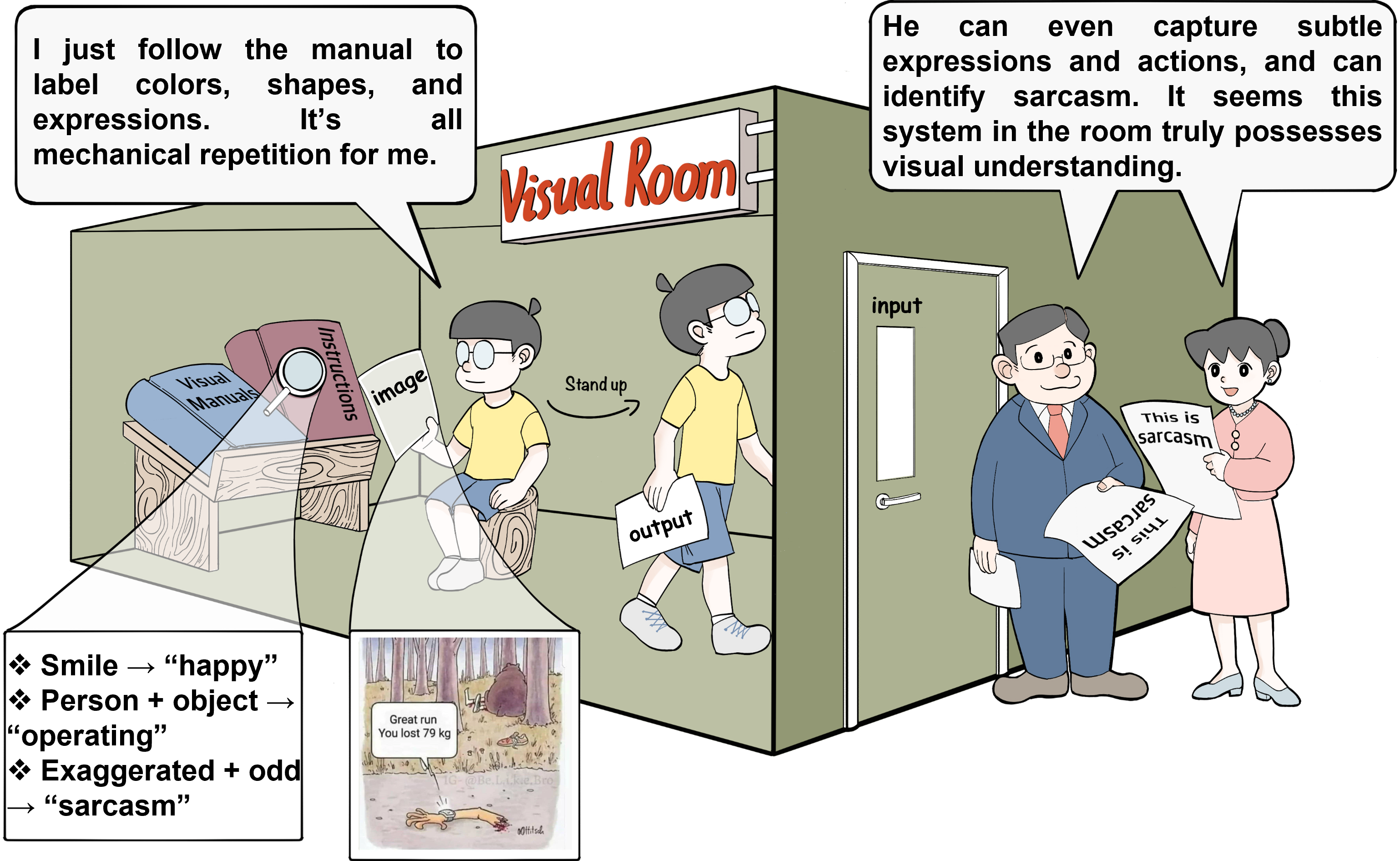}
    \caption{Visual Room}
    \label{fig:visualroom}
\end{figure}

we have proposed the Visual Room argument(VRA)\cite{mmsar},As shown in the Fig.\ref{fig:visualroom}:

\textit{Imagine a scenario where a visual symbol operator with no understanding of visual semantics is locked inside a sealed room with only a small opening. The room is filled with manuals containing rules for describing visual features such as shapes, colors, objects, and facial expressions, as well as an instruction book (e.g., a MLM) that details how to process these features. Whenever an image is passed into the room through the opening (input), the operator inside mechanically follows the instructions: (1) meticulously records all visual elements in the image, such as facial expressions, objects, and scene composition (scene recognition); (2) combines these symbolized visual details into a descriptive text, or makes a judgment: “this is sarcasm” or “this is not sarcasm” (output).  To an outside observer, this Visual Room seems capable of recognizing and describing complex images, and even correctly judging non-literal intentions such as sarcasm or metaphor. In reality, however, the operator inside has no genuine understanding of the image’s true intent, and just simply manipulating symbols according to a set of rules.}

\section{Volunteer Recruitment}
The annotation procedure consists of two phases: annotation and re-annotation. Specially, we recruit three well-educated volunteers  to take part in data annotation and re-annotation. They all signed on the consent form before the study and were paid an equal \$7.5/hour in local currency. Prior to annotation, they received professional guidance covering the use of the annotation system, the criteria for labeling, human affect-related knowledge,etc. We answered further questions from the volunteers regarding the guidance. Then, they were instructed to annotate 20 examples first to strengthen the inter-annotator agreement. After full annotation, we compute Cohen’s $\kappa$ = 0.73, indicating strong inter-annotator reliability. All questions are further reviewed by a fourth verifier.
\section{Evaluation Criteria}
We propose a perception-cognition evaluation framework to explicitly disentangle perception from cognition. S is the golden scene description for Image. $sum_{right}$ is the right number in the perception task in each examples. $sum_{error}$ is the error number in the perception task in each examples.

To ensure the robustness of short-answer question evaluations, we employ a hybrid approach combining SentenceTransformers and third-party LLM (i.e., Claude-4) judgment for model's scene description $ S' $:

\begin{equation}
  sim(S{'},S)=\frac{1}{2}*sim_{ST}(S{'},S)+\frac{1}{2}*sim_{LLM}(S{'},S)
\end{equation}
A description was considered correct if the similarity exceeded the threshold $\delta$ =0.75. 

\textbf{Perception level:} The Perception level accuracy is defined as:
\begin{equation}
  Acc_P = [\frac{sum_{right}}{sum_{right}+sum_{error}} \ge \delta]
\end{equation}

\textbf{Cognition level:} The Cognition level accuracy is defined as:
\begin{equation}
  Acc_C = P(y' = y|Acc_P = 1) 
\end{equation}
which measures the model’s cognitive capacity under correct perceptual grounding.

Hence, to quantify the gap between perception and cognition, we further define the perception-cognition gap:
\begin{equation}
  Gap_{(P,C)} = P(y' \neq y|Acc_P = 1)  
\end{equation}
\section{Model performance analysis}

\begin{figure}[htbp]
    \centering
    \includegraphics[width=1\linewidth]{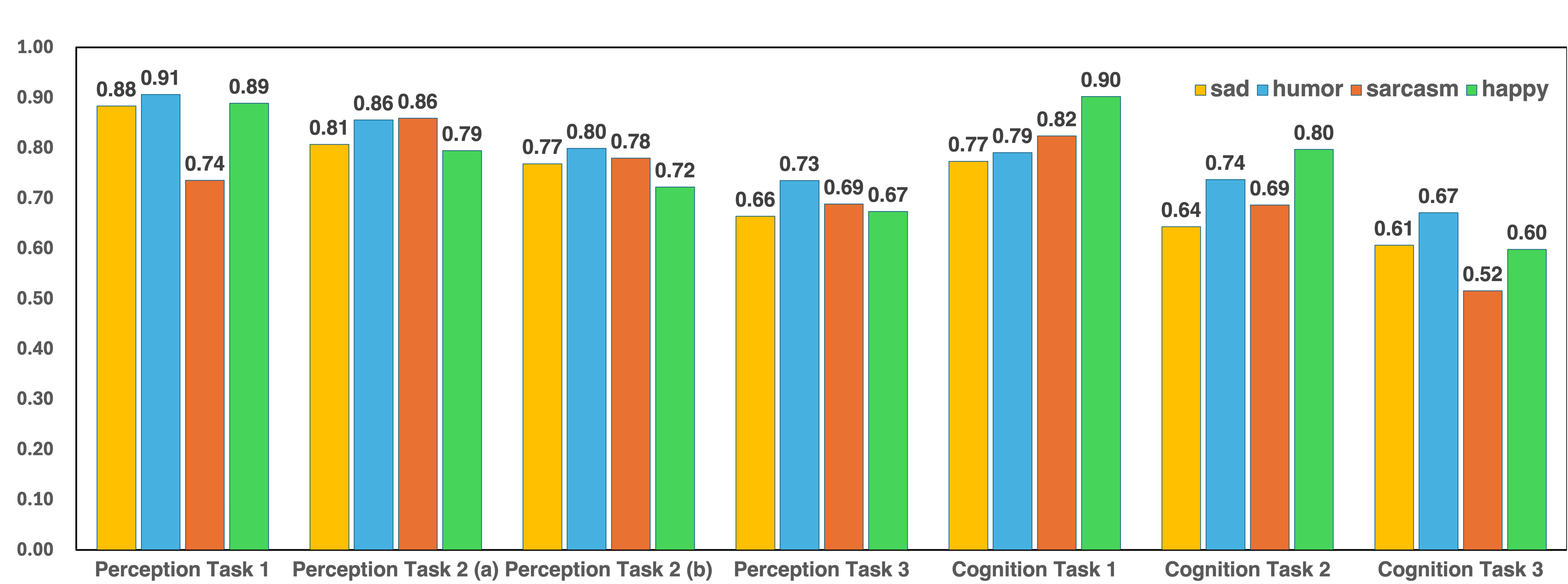}
    \caption{The average accuracy of each model in different }
    \label{main}
\end{figure}

As shown in the Fig.\ref{main}.
\textbf{GPT-5 and Gemini 2.5 Pro form the first tier}, maintaining a leading edge across most emotion and task dimensions. This suggests that foundational model capabilities, training data scale, and quality remain key determinants of performance. On relatively basic tasks like Perception 1 (Color Recognition) and Cognition 1 (Simple Inference), these top models perform stably and exceptionally well. For instance, GPT-5 achieves a remarkable 0.97 accuracy on Perception 1 for the "sad" emotion.

Meanwhile, models like \textbf{GLM-4V-Plus, GPT-o1 , Doubao and Qwen-VL-Max demonstrate strong competitiveness on specific tasks}, indicating their potential for optimization in certain capabilities. A notable example is GLM-4V-Plus achieving the highest score of 0.95 on Perception 1 for "humor," surpassing some first-tier models.

\textbf{Models with smaller parameter counts, such as Qwen2.5-VL-7B and DeepSeek-VL-7B, struggle with complex cognitive tasks.} Their scores on mid-level and high-level of cognition for "sarcasm" are significantly lower than other models (e.g., only 0.31 on Cognition 2), clearly reflecting the importance of model capacity for understanding advanced, abstract concepts.

\section{Perception-Cognition Chain Evaluation}
This study investigates the relationship between a model’s performance on perceptual tasks and its subsequent cognitive reasoning capabilities. Inspired by the Chain-of-Thought (CoT) prompting paradigm and its extensions to multimodal domains, we preserve context to see if the model can improve its performance on subsequent complex questions by answering previous simple questions. 

\begin{figure}[htbp]
    \centering
    \includegraphics[width=0.50\textwidth]{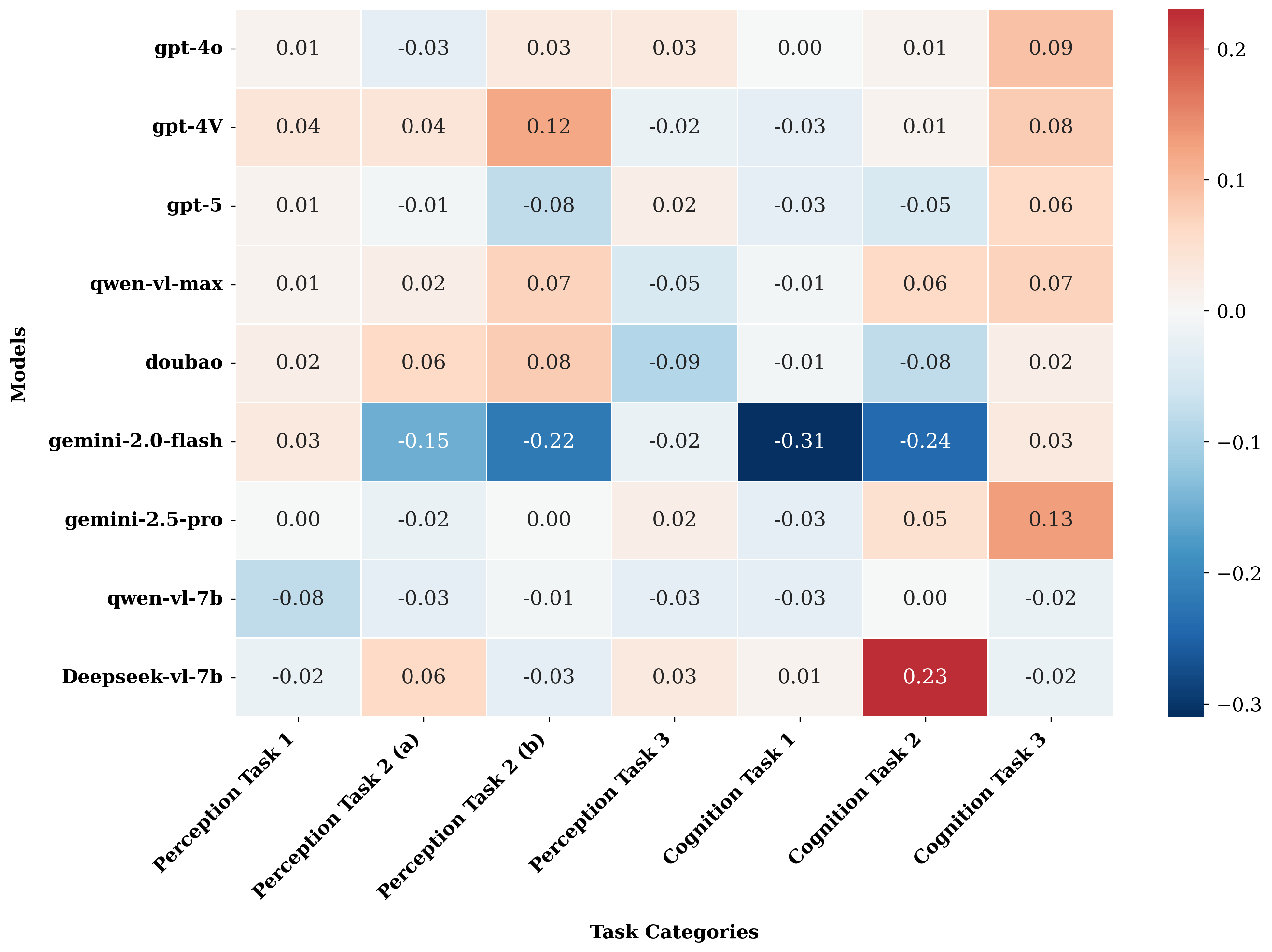}
    \caption{Heatmap after introducing Perception-Cognition Chain}
    \label{hea}
\end{figure}
As illustrated in the Fig.\ref{hea}, when queries transition from foundational knowledge to advanced tasks, most large language models (LLMs) exhibit a clear improvement in accuracy.  Analysis of the data further reveals that progressive perceptual questioning elevates the model’s cognitive comprehension of images incrementally. Notably, this approach demonstrates a more substantial enhancement in handling high-complexity cognitive tasks. While the majority of models show strengthened cognitive performance through this method, Gemini 2.0 Flash presents an opposite trend. 

The Perception-Cognition Chain evaluation reveals a nuanced picture. Providing models with progressive questions moving from basic perception to complex reasoning, improves performance on cognitive tasks for most models, with some achieving up to 15\% accuracy gains on high-level of cognition. This aligns with findings from Chain-of-Thought prompting literature, suggesting that intermediate steps can guide models toward better reasoning. However, Gemini 2.0 Flash demonstrates an inverse pattern, with performance degrading under progressive questioning. This divergence likely reflects architectural trade-offs, where optimization for reduced latency and computational efficiency compromises the model's ability to maintain and leverage extended conversational context. 

This also shows that when faced with complex reasoning tasks,  model lacks the ability to autonomously split tasks like humans. Although the introduction of guided step-by-step questioning can improve the ability of MLLMs in difficult cognitive tasks, this phenomenon also indirectly reveals the potential limitations of the current model in its autonomous reasoning ability.

\section{prompt}
\begin{lstlisting}[caption={claude 4 compare prompt}]
prompt-template = "
You are a strict similarity evaluator.
Task: Given two image descriptions, evaluate their semantic and content similarity, and output a score between 0 and 1.

Scoring criteria:
1.0 = Identical, fully interchangeable, no meaningful differences
0.8-0.99 = Highly similar, only minor wording or detail differences
0.6-0.79 = Moderately similar, main content overlaps but clear differences exist
0.4-0.59 = Partially similar, only a few elements or themes overlap
0.2-0.39 = Low similarity, barely related, minimal common points
0.0-0.19 = Completely different, no meaningful overlap

Requirements:
1. Judge strictly based on semantic content, not writing style or length.
2. Do not explain the reasoning. Output only a numeric value (decimal between 0 and 1, with two decimal places).

Input:
Description A: "{desc1}"
Description B: "{desc2}"

Output:
"
\end{lstlisting}
\end{document}